\documentclass[11pt]{article}
\usepackage[final]{acl}

\usepackage{times}
\usepackage{latexsym}
\usepackage{amsfonts} 
\usepackage{amsmath} 

\usepackage[T1]{fontenc}
\usepackage[whole]{bxcjkjatype}
\usepackage[utf8]{inputenc}

\usepackage{microtype}

\usepackage{inconsolata}

\usepackage{graphicx}
\usepackage{color}
\usepackage{multirow} 
\usepackage{subcaption}
\newcommand{\interalia}[1]{\citep[\textit{inter alia}]{#1}}
\newcommand{\todo}[1]{\textcolor{black}{#1}}

\title{Revisiting the Systematicity in Negation in the Era of In-Context Learning}

\author{Hitomi Yanaka$^{1,2,3,*}$ \and
  Taisei Yamamoto$^{1,2,*}$ \\
  $^1$The University of Tokyo, $^2$Riken, $^3$Tohoku University \\
  \texttt{\{hyanaka,yamamo96\}is.s.u-tokyo.ac.jp} \\}

\begin{document}
\maketitle
\def\thefootnote{*}\footnotetext{equal contribution.}\def\thefootnote{\arabic{footnote}}
\begin{abstract}
Understanding the meaning of negated sentences remains one of the challenges for language models, even in the era of large language models (LLMs).
We analyze systematicity regarding LLM understanding of negation from two perspectives: behavioral systematicity and representational systematicity.
For behavioral systematicity, we confirm that through demonstrations and in-context learning, LLMs can recognize negation expressions and scope within sentences to some extent, but they fail to achieve perfect performance.
In particular, the difficulty of the negation scope recognition for models varies depending on the output format.
For representational systematicity, we analyze the extent to which \todo{function vectors can be robustly constructed from in-context examples for tasks that are essential to understanding negation}.
The experiments suggest that while function vectors can be composed for negation cue extraction tasks, extracting function vectors for recognizing scope is more challenging.
\end{abstract}

\section{Introduction}
Even in the era of large language models (LLMs), language models have struggled with understanding negation \interalia{truong-etal-2023-language,garcia-ferrero-etal-2023-dataset,zhang-etal-2023-beyond}.
One reason can be the lack of \emph{systematicity}, the capacity to comprehend an unbounded set of structured expressions from a finite set of primitive components \cite{Fodor1988-FODCAC}.
In human cognition, systematicity underlies linguistic productivity: for example, understanding the sentence ``The cat didn't chase the dog'' immediately confers an ability to understand ``The dog didn't chase the cat'', due to linked representations of the constituent concepts and their combinations.

In their foundational critique of connectionist models, \citet{Fodor1988-FODCAC} argued that systematicity is a property of \emph{representations} rather than behavior: only systems that encode structured representations can exhibit systematic generalization across contexts. 
They contend that architectures lacking such representational structure fail to guarantee systematic inference, even if they occasionally produce correct outputs on specific tasks. 
This distinction highlights that evaluating model behavior on benchmark tasks is insufficient to determine whether the model has acquired the underlying representational principles that support systematic generalization.
Despite the theoretical importance of structured representations, previous model evaluation has focused on behavior alone, using monotonicity inference benchmarks involving negation that measure performance on held-out combinations of known inputs \cite{geiger-etal-2020-neural,yanaka-etal-2020-neural,goodwin-etal-2020-probing}.

Recently, \citet{vegner-etal-2025-behavioural} contrasted \emph{behavioural systematicity}, which reflects performance patterns on specific test inputs, with \emph{representational systematicity}, which concerns the internal encoding of compositional structure that supports systematic inference. 
They show that many benchmarks implicitly assume representational systematicity but only measure observable behavior, leading to ambiguous claims about models' cognitive capacities, and argue the necessity of methods that explicitly probe whether internal representations exhibit the structural regularities.
Analyzing representational systematicity is particularly salient for LLMs because they can answer unseen questions in a benchmark through spurious correlations or memorized fragments, without possessing the compositional representations that would support systematic inference \cite{kumon-yanaka-2025-analyzing}. 
Thus, there is a pressing need for representational analyses that can reveal whether LLMs encode structured elements akin to those hypothesized in cognitive theories of systematicity. 
In addition, previous research has primarily focused on analyzing the systematicity of pretrained language models such as BERT \cite{devlin-etal-2019-bert}, while the systematicity of LLMs based on in-context learning (ICL) has not been sufficiently explored.

In this paper, we analyze not only behavioral systematicity but also representational systematicity regarding LLM understanding of negation.
For behavioral systematicity, we confirm that LLMs can, to some extent, recognize negation expressions and negation scope within sentences that do not appear in the demonstration.
For the analysis of representational systematicity, we focus on the concept of function vectors \cite{todd2024function} \todo{(alongside concurrently proposed task vectors \cite{hendel-etal-2023-context})}, which are dense vectors extracted from the model's internal activations that summarize the input–output mapping implicitly demonstrated by a set of in-context examples.
Specifically, we analyze the extent to which \todo{function vectors can be robustly formed for negation understanding tasks}.

\section{Related Work}
\subsection{Function Vectors in LLMs}
\label{subsec:fv}
Recent analyses of the mechanisms underlying ICL in LLMs have examined function vectors (concurrent studies include task vectors~\cite{hendel-etal-2023-context} and in-context vectors~\cite{10.5555/3692070.3693379}), which distill underlying function information from a few input-output demonstrations into a single vector.
The vector can be extracted from the model's hidden activations and injected to trigger the same function in a different context.

\paragraph{Function vectors}
Consider an ICL prompt containing $k$ input-output demonstrations followed by a query.
When the prompt is processed by an LLM, each layer $\ell$ produces hidden states $\mathbf{h}_\ell \in \mathbb{R}^d$ at \todo{the last token position}.
The key hypothesis is that a small subset of attention heads transports a representation of the demonstrated function through the network. 
A set of attention heads that have a strong causal influence on the model's ability to perform the task can be identified by causal mediation analysis (see Appendix~\ref{app:attention_fv} for details).

Let $\mathcal{A}$ denote the set of such attention heads, each indexed by layer $\ell$ and head $j$. 
For each attention head $(\ell,j)$, the output activation for the final token of the demonstration context is extracted and averaged across multiple prompts demonstrating the same task. 
Denoting this activation by $\mathbf{a}_{\ell,j}$, the function vector is defined as the aggregated activation across the identified heads:
\[
\mathbf{v}_{\text{func}}
=
\sum_{(\ell,j)\in\mathcal{A}}
\bar{\mathbf{a}}_{\ell,j},
\]
where $\bar{\mathbf{a}}_{\ell,j}$ is the mean activation over prompts.
This vector summarizes the transformation encoded by the demonstration examples, effectively compressing the ICL task into a single vector representation. 
Empirically, these vectors tend to emerge in middle layers and remain robust across prompt variations.

After extraction, function vectors can be used to causally intervene in the model's forward pass. 
The intervention is performed by adding a function vector to the hidden state of a selected layer during inference.
Let $\mathbf{h}_\ell$ denote the hidden state at layer $\ell$ for a new input without demonstrations.
The function vector is injected via a residual intervention:
\[
\mathbf{h}_\ell' = \mathbf{h}_\ell + \mathbf{v}_{\text{func}} .
\]
The modified hidden state $\mathbf{h}_\ell'$ is then propagated through the remaining layers.
Remarkably, prior research has shown that injecting the function vector robustly induces function execution for various tasks, even when applied to zero-shot contexts that do not contain any task information.

\paragraph{\todo{Task vectors}}
Task vectors provide an alternative account of how ICL encodes task information in the hidden representations of an LLM. 
Rather than identifying a subset of causally important attention heads, task vectors are extracted directly from the residual stream activations induced by a set of demonstrations. 
The central hypothesis is that, when processing an ICL prompt, the model first computes a latent representation of the task specified by the demonstrations and then applies this representation to the query.

Consider the hidden states corresponding to a designated task-representative token position in a prompt (e.g., ``$\rightarrow$'' in an ICL setting, where input-output demonstrations are described as $\textsf{small}\rightarrow \textsf{big}$, and a target query is described as $\textsf{fast}\rightarrow$).
\citet{hendel-etal-2023-context} observed that these hidden states exhibit a common component that is largely invariant across prompts for the same task. 
They called them task vectors, which serve as compact representations of the transformation specified by the demonstrations, abstracting away from the particular examples used to define the task.
Like function vectors, task vectors can be used as interventions during inference. 
The task vector is injected \todo{by swapping the residual stream activation of the final token in the target query (``$\rightarrow$'') with that in the ICL prompt}. 
Empirical results show that the injected task vector can partially recover the behavior induced by in-context examples, suggesting that a substantial portion of the task information learned through ICL is encoded in a reusable vector representation within the model's hidden states.

\todo{If a function vector (in our paper, a function vector and task vector will be collectively referred to as a function vector) representing a task associated with arbitrary input can be constructed and it can execute the task, then LLMs} might realize a kind of representational systematicity.
However, existing research has primarily focused on tasks that can be easily inferred from demonstrations (e.g., the task to extract tokens related to animal names from a comma-separated sequence of tokens) and single-token output tasks.
In contrast, this study attempts to construct a function vector for the task of understanding the meaning of negated sentences, then analyzes the extent to which the model can perform the task on unseen negated sentences when function vectors are applied in zero-shot settings.

\subsection{Analysis of LLMs on Negation}
Negation is one of the most fundamental linguistic phenomena, which reverses the meaning of words, phrases, and sentences.
Various negation understanding benchmarks have been provided to analyze the logical reasoning abilities of language models~\cite{hossain-etal-2020-non,hossain-etal-2020-analysis,kalouli-etal-2022-negation,ravichander-etal-2022-condaqa,truong-etal-2023-language,garcia-ferrero-etal-2023-dataset,zhang-etal-2023-beyond} to evaluate the linguistic capabilities of LLMs.
Some studies of probing negation understanding have shown model insensitivity to the contextual impacts of negation~\cite{10.1162/tacl_a_00298,kassner-schutze-2020-negated,kim-etal-2025-semantic}.

At least, understanding of the functional meaning in negated sentences requires (i) the recognition of the negation expression within the sentence and (ii) the comprehension of its scope.
We analyze the systematic abilities of LLMs to handle negation by focusing on these two tasks.

\begin{table}[]
\small
\centering
\begin{tabular}{ll}\hline
\multicolumn{2}{c}{I do \texttt{[NEG]}not\texttt{[/NEG]} call customer service often.}\\ \hline
\textsc{Extract}&not call customer service often\\
\textsc{Bracket}&I do not [call customer service often].\\
\textsc{Last-word}&often\\
\textsc{NER}&$[0,0,0,1,1,1,1,1,1,1,0]$\\
\hline
\end{tabular}
\caption{Negation resolution task formats. To simplify the analysis, we tagged the target negation cues. 
}
\label{tab:taskformat}
\end{table}

\section{Behavioral Systematicity}
\subsection{Settings}
\paragraph{Overview}
We first analyze the behavioral systematicity of LLMs.
Under a few-shot learning setting, we evaluate how well LLMs can correctly recognize not only the negation expression but also its scope in input sentences composed of word combinations not appearing in the demonstrations.
To observe model behavior with simple and controllable settings, this study focuses on investigating the semantic understanding capabilities of LLMs in single sentences containing one negation expression.
We consider (i) a negation cue extraction task that extracts a negation expression in an input sentence and (ii) a negation resolution task that predicts the part of the sentence (negation scope) affected by the extracted negation cue.
\todo{As a baseline task for which function vectors can be constructed in prior research \cite{todd2024function,hendel-etal-2023-context}}, we also evaluate LLMs using the antonym prediction task.
The task is to generate a word with an opposite meaning, given an input word (e.g., input is \texttt{old:} and its expected output is \texttt{young}). 

Furthermore, various output formats are conceivable for the negation resolution task. 
If LLMs can systematically recognize the scope of negation expressions, they should robustly predict the correct answer even when the output format differs. 
Thus, we investigate how well they recognize the scope of negation expressions by varying the output format.
We consider four task formats shown in Table~\ref{tab:taskformat}: \todo{\textsc{Extract}} is an extraction format that outputs the sequence of tokens within the negation scope,
\todo{\textsc{Bracket}} is a bracket format where the negation scope is enclosed in square brackets,
\todo{\textsc{Last-word}} is a last-word format that outputs the final token of the negation scope, and
\todo{\textsc{NER}} is an NER task format that returns a sequence where tokens outside the scope of negation are 0, and tokens within the scope are 1.

\begin{figure*}
\centering
  \begin{minipage}{\linewidth}
    \centering
    \includegraphics[width=0.76\linewidth]{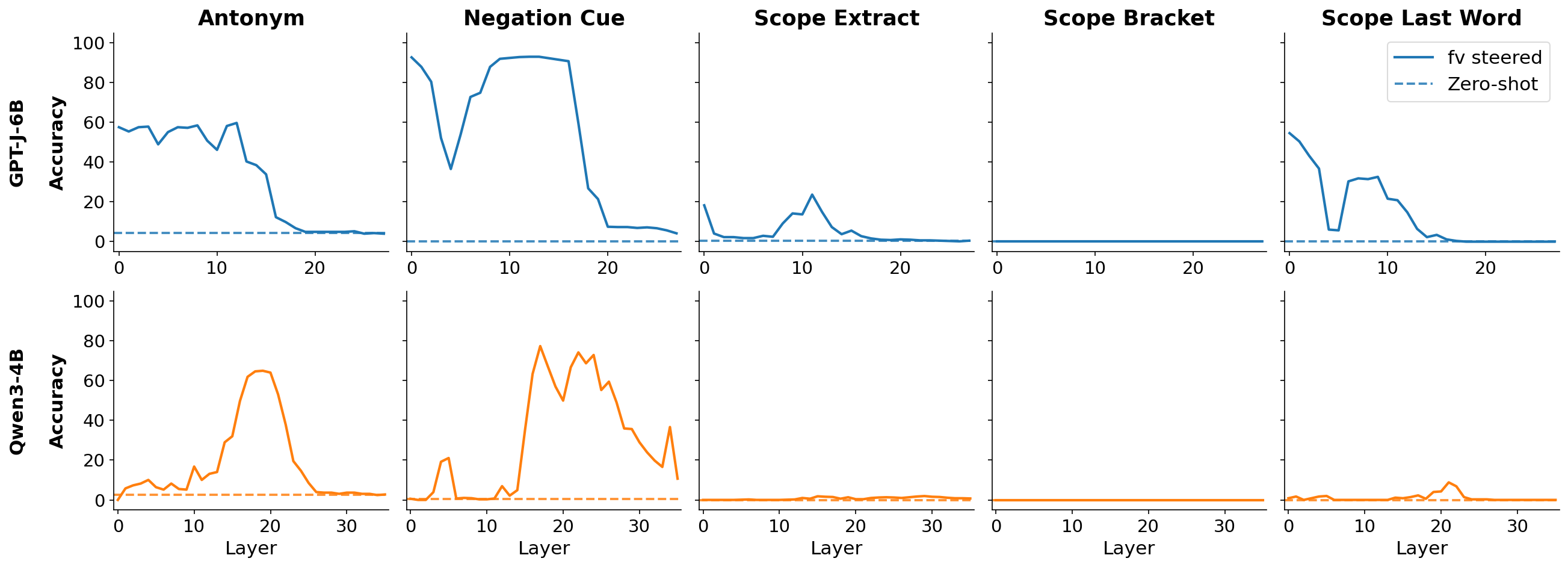}
    \subcaption{Zero-shot top-1 accuracy of applying function vectors to models.}
    \label{fig:fv}
  \end{minipage}
  \vspace{1em}
  \begin{minipage}{\linewidth}
    \centering
    \includegraphics[width=0.76\linewidth]{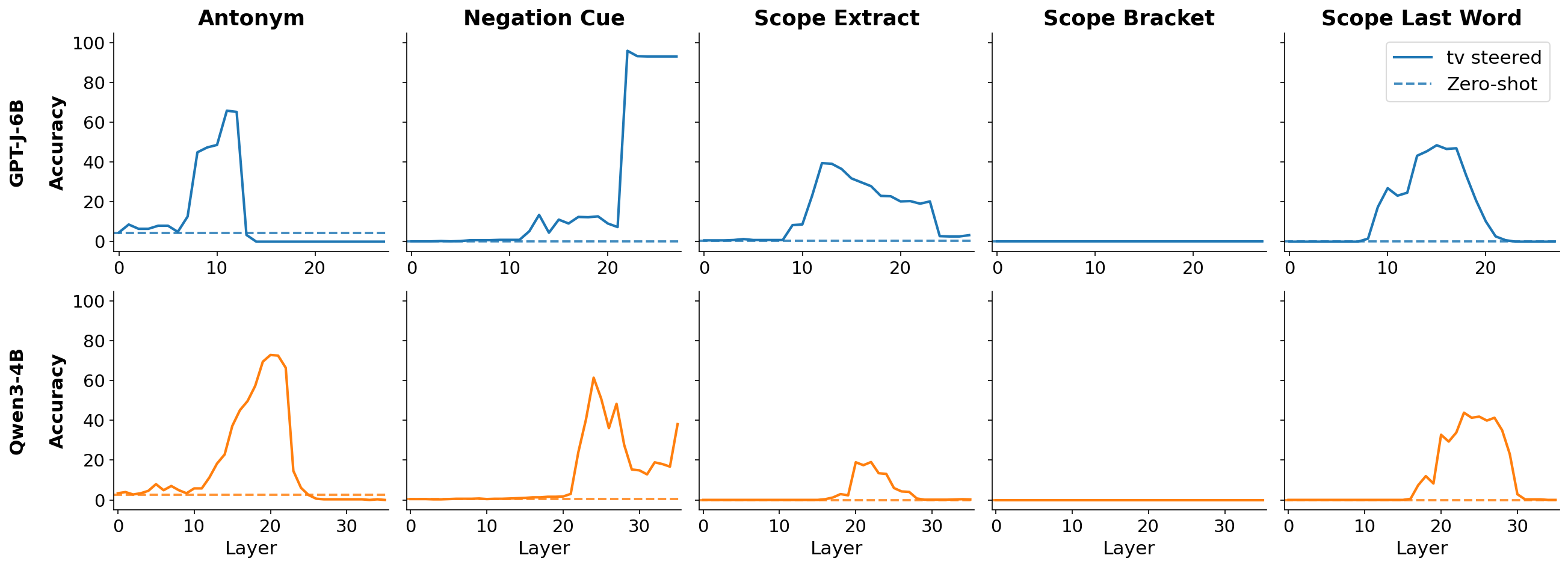}
    \subcaption{Zero-shot top-1 accuracy of applying task vectors to models.}
    \label{fig:tv}
  \end{minipage}
\caption{Representation systematicity results across tasks: antonym prediction, negation cue extraction, and negation resolution (\textsc{Extract}, \textsc{Bracket}, and \textsc{Last-word}). The dotted line indicates the accuracy in zero-shot settings. \todo{Since we use test examples for which the model answers correctly in 10-shot settings, the accuracy in 10-shot settings is 100\% in this experiment.}}
\label{fig:layer}
\end{figure*}

\paragraph{Dataset}
We used SFU Review Corpus~\cite{konstantinova-etal-2012-review}, which consists of 400 real-world documents of various genres manually annotated at the token level with negation expressions and at the sentence level with their linguistic scope.
For simplicity of analysis, we sampled data from the corpus that contained at least one negation expression (\textit{not}, \textit{no}, or \textit{n't}) followed by a scope. The resulting dataset is composed of 153 training examples, 308 development examples, and 1078 test examples.
For the antonym prediction task, we use the dataset developed by \citet{todd2024function}, composed of 700 training examples, 90 development examples, and 210 test examples.
We have confirmed that the sentences in training, development, and test sets are composed of different word combinations, respectively.
For each task, we sample 10-shot in-context examples from the training set and prepend them to the test example for evaluation.
Note that, since during pre-training LLMs may learn lexical items and syntactic structures that are subsequently questioned in the test set, rigorously evaluating the systematicity of LLMs is difficult.
Thus, we relax the definition of systematicity; we analyze whether models obtain the ability to systematically capture the task from demonstrations without any instruction and correctly execute the task even for negated sentences containing lexical combinations not appearing in the demonstrations.

\paragraph{Models}
Following the prior research, we selected GPT-j-6B~\cite{gpt-j}. 
We also used a relatively recent LLM, Qwen3-4B~\cite{qwen3technicalreport}.
We compared some baselines to verify whether models merely capture heuristics rather than understanding negation (see Appendix~\ref{sec:appendix}).

\subsection{Results}
Table~\ref{tab:bahavior} shows accuracy for each task and format.
As a general trend, Qwen3-4B performed better than GPT-j-6B.
For the negation cue extraction task, both models achieved an accuracy comparable to that of the antonym prediction task. 
However, since this accuracy is not perfect, it suggests that they fail to acquire the systematic ability to recognize negation expressions for arbitrary negated sentences from demonstrations.
For the negation resolution task, the model accuracies varied depending on the task format; models especially struggle with learning the NER format from demonstrations.

\begin{table}[]
\small
\centering
\begin{tabular}{c|l|lll}\hline
\multicolumn{2}{l|}{Task}                   & GPT-j-6B & Qwen3-4B & Base.\\ \hline
\multicolumn{2}{l|}{Antonym}                & 64.5 & 65.1 & -   \\ \hline
\multicolumn{2}{l|}{Neg-cue}           & 62.0 & 64.3 &  64.7  \\ \hline
\multirow{4}{*}{Scope} &\textsc{Extract}&  56.9 & 76.0 & 55.7       \\
                                &\textsc{Bracket}& 22.6 & 69.8 & 55.7   \\
                                &\textsc{Last-word}& 24.5 & 32.6 & 56.9 \\
                                &\textsc{NER}& 0.0 & 0.0 & 55.7 \\ \hline     
\end{tabular}
\caption{Model accuracy (\%) for each task and format \todo{on the 10-shot setting}.
}
\label{tab:bahavior}
\end{table}

\section{Representational Systematicity}
\subsection{Settings}
Next, we investigate the representational systematicity of LLMs.
We analyze the extent to which applying a function vector/task vector to each layer at the final token position of the prompt can cause the model to perform a task in contexts that differ from the ICL contexts from which it was extracted.
\todo{For each task, we compute a function vector~\cite{todd2024function} and a task vector~\cite{hendel-etal-2023-context} respectively, by using 10-shot demonstrations sampled from training examples.}
\todo{As for a function vector setting, if the output is composed of multiple tokens, the function vector is added to all tokens following the final token.}

According to \citet{todd2024function}'s evaluation setting, we only include test examples for which the model answers correctly given 10-shot in-context examples; we calculate accuracy on this filtered test set as the proportion of the model's task performance encoded by function/task vectors.
We used the same datasets and models, and the NER format was excluded from the analysis due to the low model performance on the 10-shot setting.

\subsection{Results}
Figure~\ref{fig:fv} and Figure~\ref{fig:tv} show zero-shot top-1 accuracy of applying function vectors and task vectors to models across tasks, respectively.
Regarding the negation cue extraction task, applying function vectors in middle layers and task vectors in late layers seems to have a large effect \todo{(90\% accuracy for GPT-J-6B and 80\% for Qwen3-4B)}.
This indicates that a faithful function/task vector is formed for the negation cue task, similar to the antonym task.

In contrast, for the negation resolution task, the function/task vector seems to have less effect \todo{(0\%-50\% accuracy)} than that for the negation cue task.
\todo{With GPT-j-6B, there is a slight effect of steering in the \textsc{Extract} and \textsc{Last-word} formats, but it is minimal.}
This suggests that constructing function/task vectors is more challenging for the negation resolution task than for the negation cue extraction task.
There are two possible reasons: (i) the task of extracting multiple tokens from sentences might be difficult to recognize, and (ii) models still struggle with capturing the concept of negation scope from in-context examples.
Further investigation should reveal the bottlenecks in the LLM's systematic recognition of negation scope.

\todo{Regarding the differences between function vectors and task vectors, Figure~\ref{fig:fv} and Figure~\ref{fig:tv} suggest that task vectors can be more successfully used as interventions than function vectors in negation resolution tasks across models.
In addition, task vectors tend to be more effective than function vectors at later layers.
While function vectors are constructed using a small number of key attention heads from middle and late layers (see Appendix \ref{app:attention_fv}) and represent input-output functions, task vectors are constructed from a residual stream and are thought to contain more global and abstract task information.
Since later layers handle more abstract representations~\cite{xu2026emergentstructuredrepresentationssupport}, task vectors can execute tasks more effectively at later layers.
Another possible cause might be intervention differences between function vector addition and task vector replacement.
Adding function vectors in later layers cannot correct information that has already been computed in previous layers.
In contrast, task vectors replace the activation in the residual stream, making it possible to overwrite past information, including earlier computational results.
Further analysis is needed using a wider variety of controlled tasks and formats.}

\section{Conclusion}
We investigated LLM understanding of negation from two perspectives: behavioral systematicity and representational systematicity.
For behavioral systematicity, experiments showed that through demonstrations and ICL, LLMs can recognize negation expressions and negation scope within sentences involving different word combinations to some extent, but their performance is not perfect.
In particular, the difficulty of the negation resolution task for models varies depending on the output format.
For representational systematicity, we analyzed \todo{the extent to which function vectors necessary for understanding negation can be robustly constructed from in-context examples}.
The experiments suggest that, \todo{although the function vectors for negation cue extraction can be formed, extracting the function vector for scope recognition is more challenging}.
Our analysis provides insights into the bottlenecks in current LLMs' systematic understanding of negation from behavior and representation perspectives, hopefully leading to improvements in LLMs' understanding of negation.

\section*{Limitations}
Since LLMs may have encountered sentences from the dataset used in this study (the SFU review corpus) during pre-training or other processes, it is difficult to rigorously analyze their systematic abilities when provided with sentences involving unseen word combinations.
In this study, we have adopted a slightly broader definition of systematicity and focused our analysis on the systematic abilities to capture the task from demonstrations without any instruction and execute the task for sentences involving word combinations not seen in the demonstrations.

Another limitation is that for simplicity of analysis, our data is limited to sentences involving at least one negation expression (\textit{not}, \textit{no}, or \textit{n't}) followed by a scope.
However, in real texts, there are various negation expressions, and sometimes the negation scope precedes the negation expression (e.g., \textit{To Luke ten years in age difference is nothing} from the SFU review corpus).
Also, as briefly introduced in Section~\ref{subsec:fv}, \todo{several function vector techniques have been explored.
In future work, we will consider multiple function vector techniques} and analyze models with sentences containing diverse negative expressions and scope patterns.

Lastly, although this study focuses on the model analysis in the systematic nature of negation, other systematic linguistic and inferential phenomena could also be considered.
The proposed analysis framework can be extended to address polarity determination regarding monotonicity inference~\cite{hu-moss-2018-polarity,icard-iii-moss-2014-recent}, enabling further analysis of LLMs.
We believe that the direction of this analysis can deepen our understanding of how LLMs generalize, where they fall short, and how hidden-state interventions might yield representations that more closely resemble the compositional structure of human cognition.

\section*{Acknowledgements}
We thank the anonymous reviewers for their helpful comments and constructive feedback.
This work was partially supported by JST CREST Grant Number JPMJCR2565, JST BOOST Program Grant Number JPMJBY24H5, and  JSPS KAKENHI Grant Number JP24H00809, Japan.

\bibliography{custom}

\appendix

\section{Baseline Setting}
\label{sec:appendix}
We prepared a baseline for each task.
For the negation cue task, we calculated the accuracy using the negation cue \textit{not} as the baseline.
For the negation resolution task (\textsc{Extract}, \textsc{Bracket}, and \textsc{NER} formats),
we calculated the accuracy using the negation cue and all subsequent tokens as the baseline.
For the \textsc{Last-word} format, we calculated the accuracy using the last word of input as the baseline.

\begin{figure*}[t]
\centering
\includegraphics[width=0.76\linewidth]{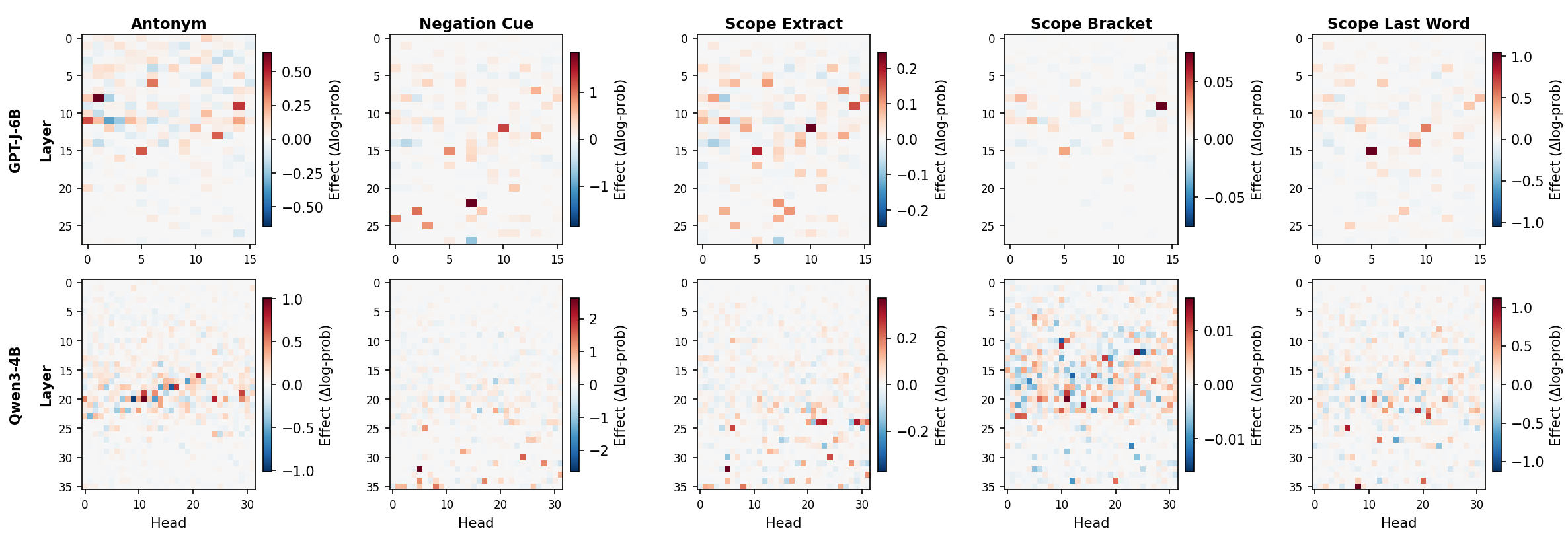}
\caption{Average indirect effect across tasks for each attention head.}
\label{fig:he_heatmap}
\end{figure*}

\section{Identifying Attention Heads for Function Vectors}
\label{app:attention_fv}
Following \citet{todd2024function}, we identify a set of attention heads that exhibit strong causal effects on the model's ability to perform the task.
To estimate the causal effect of each head, we replace its output with $\bar{\mathbf{a}}_{\ell,j}$, the mean activation of that head computed over the development set prompts, and evaluate the model's zero-shot performance on the development set.
The causal effect is defined as the average increase in the probability assigned to the correct tokens relative to the base mode (i.e., without replacement).
We then select the top ten heads to construct the function vector.

Figure \ref{fig:he_heatmap} shows the average indirect effect across tasks for each attention head in GPT-j-6B and Qwen3-4B.
We can see that the attention heads with high average indirect effect appear from the middle layers to the late layers.

\end{document}